\pdfoutput=1
\documentclass[11pt]{article}
\usepackage{tabularx}

\usepackage[final]{acl}
\usepackage{flushend}

\usepackage{times}
\usepackage{latexsym}
\usepackage{amsmath}

\usepackage[T1]{fontenc}

\usepackage{pdfpages}
\usepackage{multirow}
\usepackage{adjustbox}
\usepackage{subfigure}
\usepackage{amssymb}% http://ctan.org/pkg/amssymb
\usepackage{booktabs}
\usepackage{pifont}
\usepackage{newtxtext}
\usepackage{enumitem}
\usepackage{algorithm}
\usepackage{hyperref}
\usepackage{algpseudocode}
\usepackage{amsmath}
\usepackage{semantic}
\usepackage{newtxtext}
\usepackage{color}

\usepackage[utf8]{inputenc}

\usepackage{microtype}

\usepackage{inconsolata}

\usepackage[utf8]{inputenc}
\usepackage{graphicx}

\title{\textsc{Gurukul AI} : An Interactive AI-Driven Educational Platform for Indian Education System}

\author{
 \textbf{Isha Narang\textsuperscript{$\dagger$}},
 \textbf{Sneh Gosai\textsuperscript{$\ast$}},
 \textbf{Mayank Singh\textsuperscript{$\dagger$}}
\\
\\
 \textsuperscript{$\dagger$}Indian Institute of Technology Gandhinagar
\\
 \textsuperscript{$\ast$}Pandit Deendayal Energy University
\\
 \small{
   \textbf{Correspondence:} \href{mailto:singh.mayank@iitgn.ac.in}{singh.mayank@iitgn.ac.in}
 }
}

\begin{document}

\maketitle

\begin{abstract}
Recent advances in large language models (LLMs) like ChatGPT and LLaMA have transformed AI-driven education, but these systems are predominantly trained on Western-centric data, making them ill-suited for regional curricula like India’s. The Indian education system is linguistically diverse, exam-oriented, and structured around standardized syllabi not addressed by existing datasets or tools. In this work, we curate a syllabus-aligned QA dataset based on NCERT (National Council of Educational Research and Training) textbooks for classes 9-12, capturing the content, context, and teaching style of Indian curricula. The final dataset, comprising 18,720 question–answer pairs across five subjects, is publicly available at \url{https://huggingface.co/datasets/LingoIITGN/Gurukul}. We fine-tune the LLaMA 3.1 8B model using this dataset and deploy it in a Retrieval-Augmented Generation (RAG) framework tailored to educational needs. We introduce \textbf{GurukulAI}\footnote{The code for the hosted website is available at \url{https://github.com/lingo-iitgn/GurukulAI}.}, an open-access platform that enables Indian students to chat with the model, get doubts cleared, practice exam-style questions, receive contextual answers, and interact in both English and Hindi. By localizing AI for Indian classrooms, our work bridges the gap between global LLM capabilities and regional educational demands. 
%The platform is publicly accessible at: \url{https://lingo.iitgn.ac.in/gurukulai/}.
\end{abstract}

\section{Introduction}
Large language models (LLMs) such as ChatGPT and LLaMA have enabled new modalities of student interaction through conversational tutoring, automated feedback, and curriculum-aligned assistance \cite{wang2023survey, khan2023survey}. However, these systems are predominantly trained on Western academic content, limiting their effectiveness for linguistically diverse, exam-driven contexts like India \cite{gilardi2023chatgpt, kakwani2020indicnlp}.

India’s education system serves over 248 million students across CBSE, ICSE, and 25+ state boards \cite{udise2023}. Most follow a syllabus-centric structure grounded in NCERT textbooks and demand explainable, multilingual, and curriculum-aligned educational tools.

Existing QA datasets such as \textsc{SQuAD} \cite{rajpurkar2016squad} and \textsc{HotpotQA} \cite{yang2018hotpotqa} are domain-generic and English-only, while Indian datasets like \textsc{iExamQA} \cite{garg2023iexamqa} lack textbook alignment. Indian EdTech platforms offer retrieval-based solutions but do not provide open, generative QA grounded in textbooks.

We introduce \textsc{GurukulAI}, a syllabus-aligned AI system tailored for Indian school education. Built on a novel dataset of 18,720 question–answer pairs derived from NCERT textbooks (Classes 9–12), the system combines a fine-tuned LLaMA 3.1 model with retrieval-augmented generation (RAG), and supports both English and Hindi subjects.

In summary, our contributions are:

\begin{itemize}
    \item \textbf{Curriculum-Aligned QA Dataset:} A novel dataset of 18,720 bilingual Q\&A pairs across five subjects (Classes 9–12), annotated with chapter-specific metadata.
    \item \textbf{Open-Access Educational Platform:} A deployed learning portal with generative tutoring, RAG-based inference, image-based question handling, and MCQ/theory modes.
    \item \textbf{Robust Evaluation:} Quantitative evaluation on 700 test questions and qualitative feedback from 40 students confirm educational alignment and usability.
\end{itemize}

\section{Related Work}

Artificial intelligence (AI) has significantly impacted education by enabling tools for personalized tutoring, formative feedback, and adaptive learning experiences. Recent surveys~\cite{wang2023llms,khan2023chatgpt,holstein2018classroom} highlight the growing role of large language models (LLMs) in educational technologies. However, many of these models are predominantly trained on Western-centric content, limiting their generalization and cultural relevance in linguistically diverse, exam-oriented educational systems like India~\cite{gilardi2023nonwestern,kakwani2020indic}.

\subsection{AI Tools in Global Education}
Global AI-powered educational platforms such as \textit{EduChat}~\footnote{\url{https://educhat.one}} and \textit{QANDA}~\footnote{\url{https://qanda.ai}} illustrate the potential of LLMs at scale. EduChat provides personalized essay feedback and emotional support using conversational agents, while QANDA leverages OCR and retrieval to solve mathematical problems from images. While these systems demonstrate strong technical capabilities, they lack alignment with national curricula such as India’s NCERT and often do not support bilingual interaction, textbook-grounded responses, or culturally relevant reasoning.

\subsection{Indian EdTech Efforts}
Several Indian initiatives such as \textit{Class Saathi}~\footnote{\url{https://class-saathi.web.app}}, \textit{Sampark Foundation}~\footnote{\url{https://www.samparkfoundation.org}}, and \textit{Minecraft Education Edition}~\footnote{\url{https://education.minecraft.net}} target scalable education solutions. Class Saathi integrates clicker-based student assessments with backend analytics, and Sampark Foundation focuses on rural learning through audio-visual tools in regional languages. However, these efforts are primarily oriented toward engagement and evaluation rather than open-domain generative question answering or curriculum-grounded learning. They lack capabilities like subject-specific reasoning or adaptive question generation that modern LLM-based systems can enable.

\subsection{Datasets for Educational NLP}
Multiple QA datasets have been developed globally to benchmark reading comprehension and reasoning. Standard datasets such as SQuAD~\cite{rajpurkar2016squad}, HotpotQA~\cite{yang2018hotpotqa}, DROP~\cite{dua2019drop}, EdNet~\cite{choi2020ednet}, and RACE~\cite{lai2017race} are valuable benchmarks, but are largely domain-general, monolingual (English), and not aligned with school-level syllabi.

In the Indian context, efforts such as iExamQA~\cite{garg2023iexamqa} and the IndicNLP Suite~\cite{kakwani2020indic} have enabled multilingual and multimodal benchmarks. More recently, IndicQA by AI4Bharat~\footnote{\url{https://huggingface.co/datasets/ai4bharat/IndicQA}} and the INDIC QA Benchmark~\cite{singh2025indicqa} offer multilingual evaluation datasets across multiple Indian languages. However, these resources still lack explicit chapter-wise alignment with NCERT textbooks or subject-aware QA capabilities that mirror classroom learning.

\subsection{Gap Addressed by GurukulAI}
To the best of our knowledge, no prior dataset offers large-scale, syllabus-aligned, chapter-tagged question–answer pairs across multiple subjects in Indian school education. Furthermore, existing generative QA systems do not integrate multimodal features and textbook-level retrieval within a unified platform. GurukulAI fills this gap by combining a fine-tuned LLaMA model with a retrieval-augmented inference engine grounded in NCERT content, enabling interactive, syllabus-aware tutoring for Classes 9–12.

\section{Data Curation}

To support the development of \textsc{GurukulAI}, we curated a syllabus-aligned question–answer (Q\&A) dataset grounded in NCERT textbooks for Classes 9 to 12. The dataset comprises 18,720 Q\&A pairs across five subjects: Science, Mathematics, Social Science, English, and Hindi. To our knowledge, this is the first publicly available QA dataset aligned with Indian school curricula, featuring chapter-level annotations and subject-wise tagging in both English and Hindi.

\subsection{Data Collection and Processing}

We collected content from three primary sources: (1) official NCERT PDFs in English and Hindi, (2) previous year CBSE question papers from public repositories, and (3) educational websites such as Vedantu\footnote{\url{https://www.vedantu.com}}, Byju’s\footnote{\url{https://byjus.com}}, LearnCBSE\footnote{\url{https://www.learncbse.in}}, and Tiwari Academy\footnote{\url{https://www.tiwariacademy.com}}. NCERT content was parsed using \textbf{PyMuPDF (Fitz)}~\cite{pymupdf2022} for layout-preserving extraction, while dynamic websites were scraped using a mix of manual curation and \texttt{Selenium}-based automation~\cite{selenium}.

\begin{table}[h]
\centering
\begin{tabular}{lcc}
\hline
\textbf{Subject} & \textbf{Classes} & \textbf{Q\&A Pairs} \\
\hline
Science & 9–12 & 8,213 \\
Mathematics & 9–12 & 4,824 \\
Social Science & 9–12 & 2,647 \\
English & 9–12 & 1,998 \\
Hindi & 9–12 & 1,038 \\
\hline
\textbf{Total} & & \textbf{18,720} \\
\hline
\end{tabular}
\caption{Distribution of Q\&A pairs across subjects.}
\label{tab:dataset-summary}
\end{table}

Each extracted item was tagged with its class, subject, and chapter number. We used the Claude Sonnet API~\cite{anthropic2023claude} for intelligent filtering and refinement. Specifically, the model was prompted to:

   \begin{figure*}[ht]
\centering

\subfigure[Chat portal]{
    \includegraphics[width=0.45\textwidth]{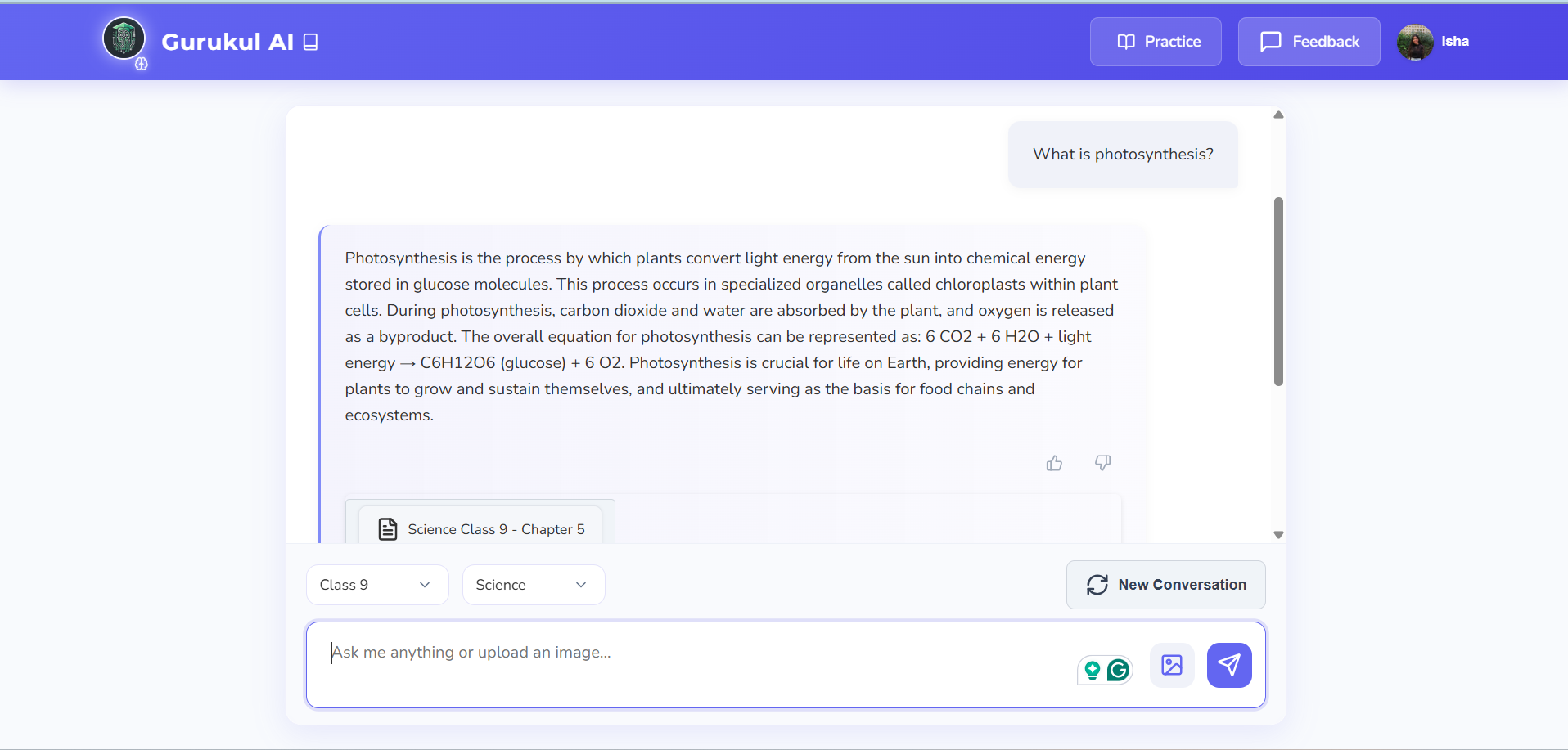}
    \label{fig:portal}
}
\hspace{0.02\textwidth}
\subfigure[MCQ Practice Portal]{
    \includegraphics[width=0.48\textwidth]{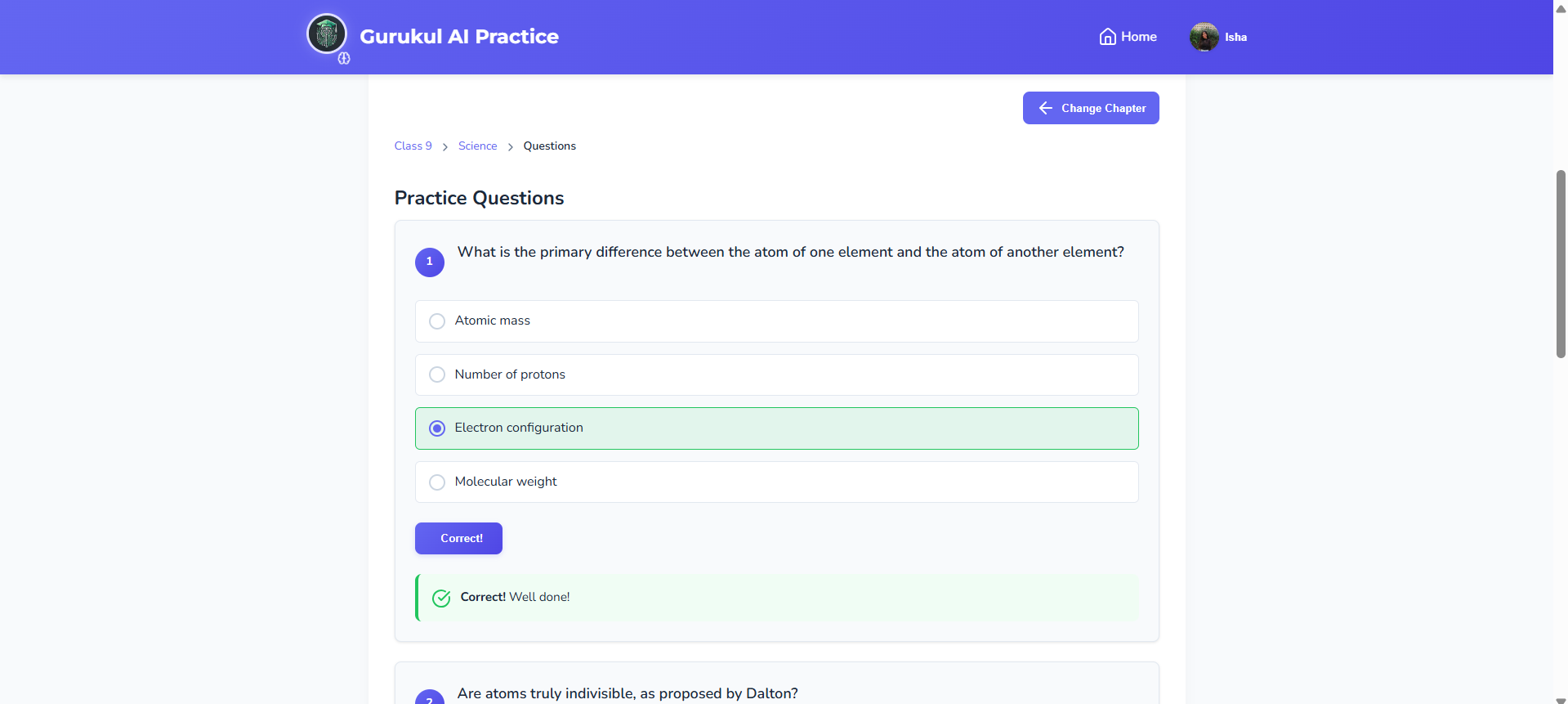}
    \label{fig:practice2}
}

\caption{Screenshots of the Gurukul AI Platform.}
\label{fig:screenshots}
\end{figure*}

\begin{itemize}
    \item Remove incomplete or visually dependent Q\&As (e.g., diagram-based).
    \item Rewrite ambiguous or compound questions into self-contained forms.
    \item Fill missing components and align phrasing with textbook tone.
\end{itemize}
Final outputs were reviewed manually for fluency, syllabus relevance, and coherence. Entries requiring visual context were excluded to retain a purely text-based dataset.

Table~\ref{tab:dataset-summary} presents the subject-wise distribution of the curated Q\&A pairs. The dataset is publicly available at: \url{https://huggingface.co/datasets/LingoIITGN/Gurukul}.

\section{GurukulAI: System Design and Implementation}

\subsection{System Architecture}

\textsc{GurukulAI} adopts a modular pipeline comprising three core components: (1) A fine-tuned generative language model, (2) A retrieval-augmented generation (RAG) module, and (3) A user-facing web interface. This architecture enables grounded, syllabus-aligned QA with support for bilingual interaction.

We selected the open-source LLaMA 3.1 8B model for its strong performance on Indic languages and availability for fine-tuning. The model was fine-tuned using Low-Rank Adaptation (LoRA) with instruction-style prompts based on our curated NCERT-aligned dataset. Training was conducted on an NVIDIA A100 40GB GPU with a context window of 4,096 tokens. The objective was to adapt the base model for textbook-aware answering across subjects while preserving general language fluency.

\begin{figure}[ht]
 \centering
 \includegraphics[width=\linewidth, trim={50 100 180 120}, clip]{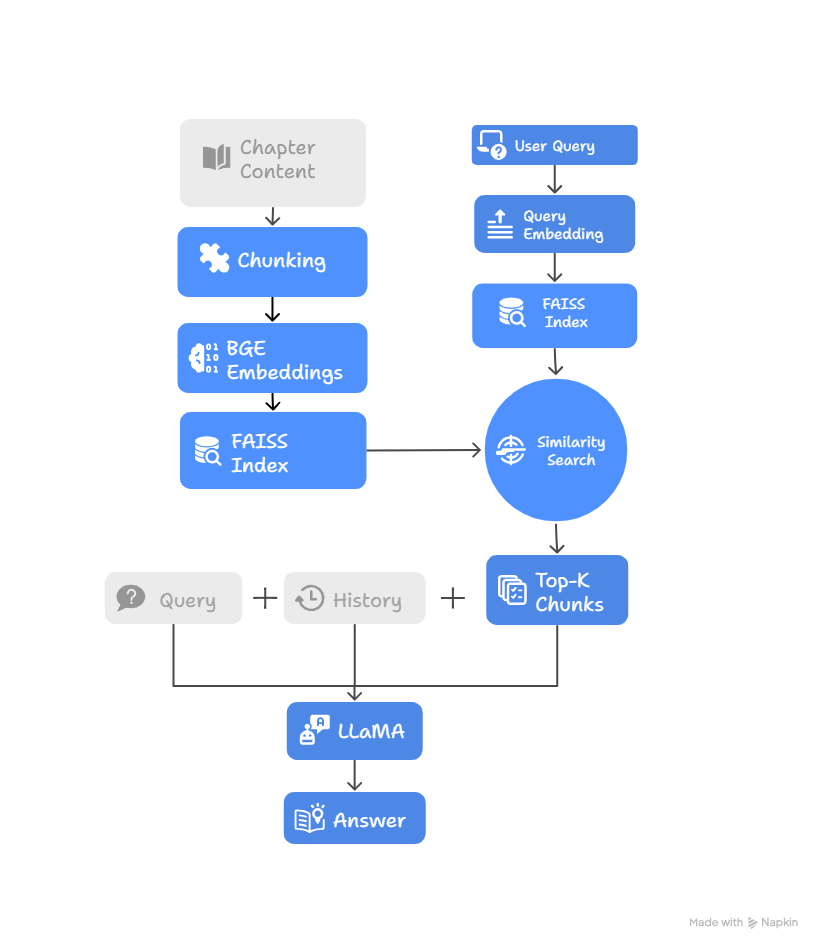}
\caption{System architecture of GurukulAI. }

 \label{fig:rag}
 \end{figure}
 
For grounding and retrieval, NCERT chapters were chunked into semantically coherent passages ($\sim$4,000 characters) using a sliding window and paragraph boundary heuristics. These passages were embedded using the BGE-m3 model~\cite{bge2023} and indexed with FAISS~\cite{faiss2017}. At inference, user queries are encoded, top-5 relevant chunks are retrieved, and combined with the original query to form a context-rich prompt passed to the LLM.

To minimize response latency, we implemented KV (key-value) caching at inference and query deduplication to avoid repeated retrieval for similar inputs. Prompt templates are dynamically generated per query to format the retrieved context and maintain instruction-following behavior.

The system is deployed on a backend server with GPU acceleration and exposed via a web interface built with Flask.  The frontend supports both English and Hindi input, with integrated modes for chatting, MCQ solving, and theory question practice.

Figure~\ref{fig:rag} shows the inference pipeline of \textsc{GurukulAI}, where a user query is embedded using BGE-m3, top passages are retrieved via FAISS from the NCERT corpus, and a context-rich prompt is formed and fed to a fine-tuned LLaMA model for generation. This RAG design ensures textbook-aligned and context-aware answers.

\subsection{AI-Enhanced User-Facing Features}

The publicly accessible \textit{GurukulAI} platform supports several interactive learning features:

\begin{itemize}
    \item \textbf{Image-Based Doubt Resolution:} Users can upload scanned or handwritten questions. For digital PDFs, text is extracted using \texttt{PyMuPDF (Fitz)}.

    \item \textbf{Follow-Up Questions and Learning Paths:} The LLaMA model suggests three contextually related questions after each interaction to promote conceptual continuity and revision.

    \item \textbf{Source Linking:} Each answer is grounded in the most relevant NCERT chapter, selected via semantic retrieval during the inference process.

    \item \textbf{MCQ Practice Mode:} After selecting a chapter, MCQs are generated using prompt-based inference over chapter summaries. Each option is explained and scored in real time.

    \item \textbf{Theoretical Answer Practice:} The model generates descriptive questions from selected chapter content and evaluates free-form student responses, offering feedback and improvement suggestions.
\end{itemize}

Figure~\ref{fig:screenshots} shows screenshots of the GurukulAI chat portal and the MCQ practice interface.

\section{Evaluation}

We evaluate \textit{GurukulAI} across three axes: (1) automatic evaluation of generated answers, (2) qualitative user study with students, and (3) ablation testing of core components like retrieval.

\subsection{Automatic Evaluation}

We assess system performance on a held-out test set of 700 questions across four subjects (Science, Social Science, English, and Hindi), with difficulty levels labeled as easy, medium, and hard. Ground truth answers were manually curated from NCERT textbooks.
\begin{figure*}[ht]
    \centering

    \subfigure[By Subject]{
        \includegraphics[width=0.45\textwidth]{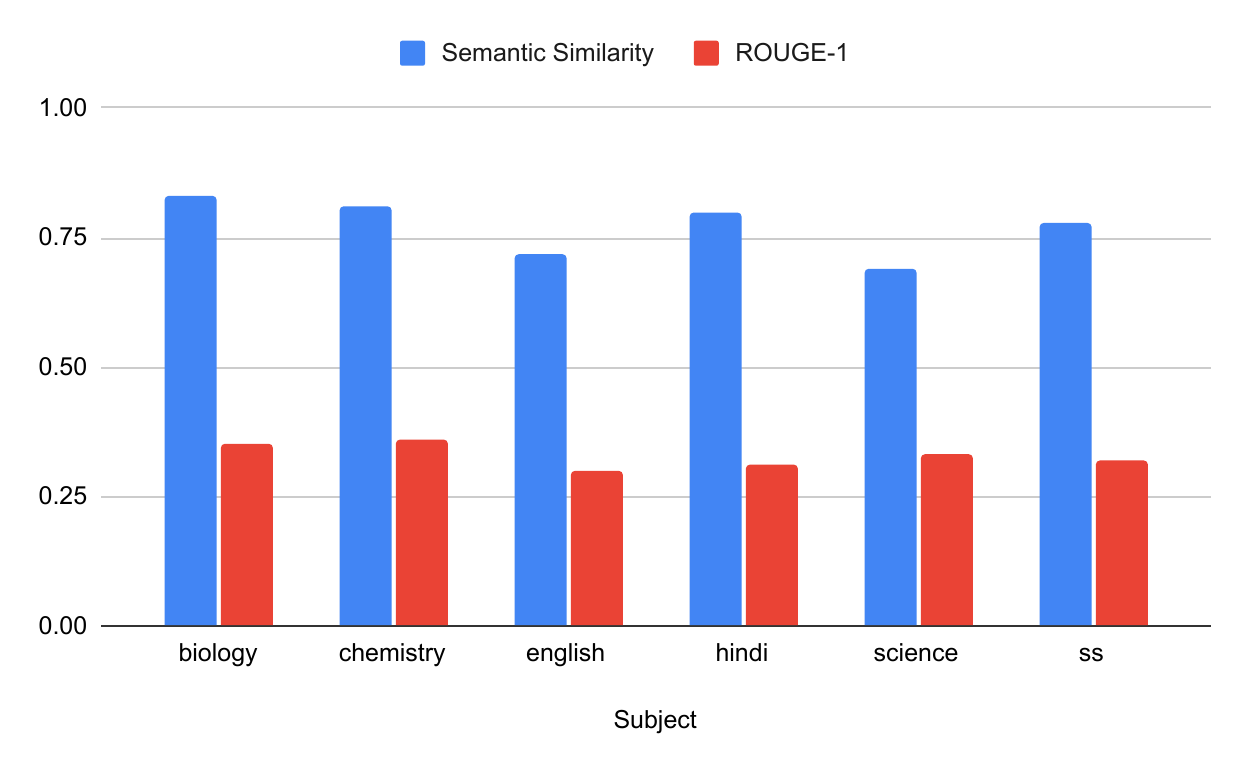}
        \label{fig:perf-subject}
    }
    \hfill
    \subfigure[By Difficulty]{
        \includegraphics[width=0.45\textwidth]{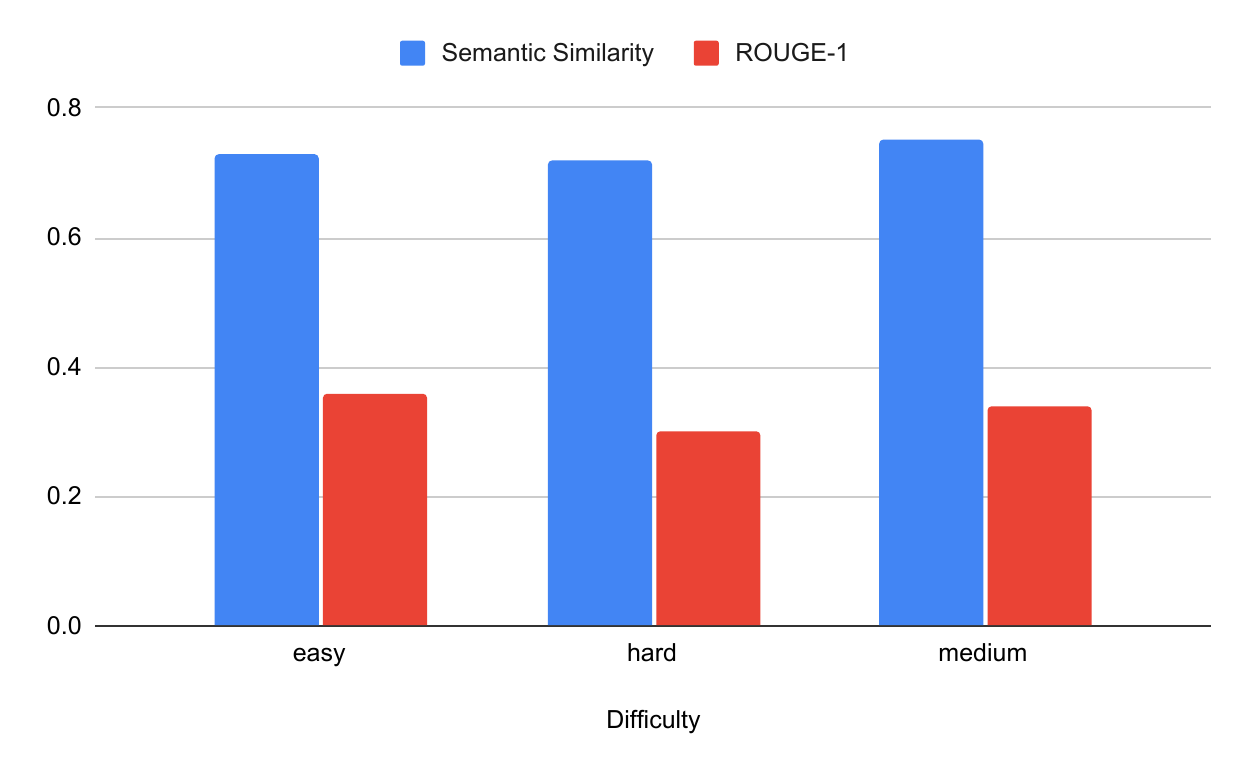}
        \label{fig:perf-difficulty}
    }

    \vspace{0.5em} % optional vertical spacing

    \subfigure[By Class]{
        \includegraphics[width=0.45\textwidth]{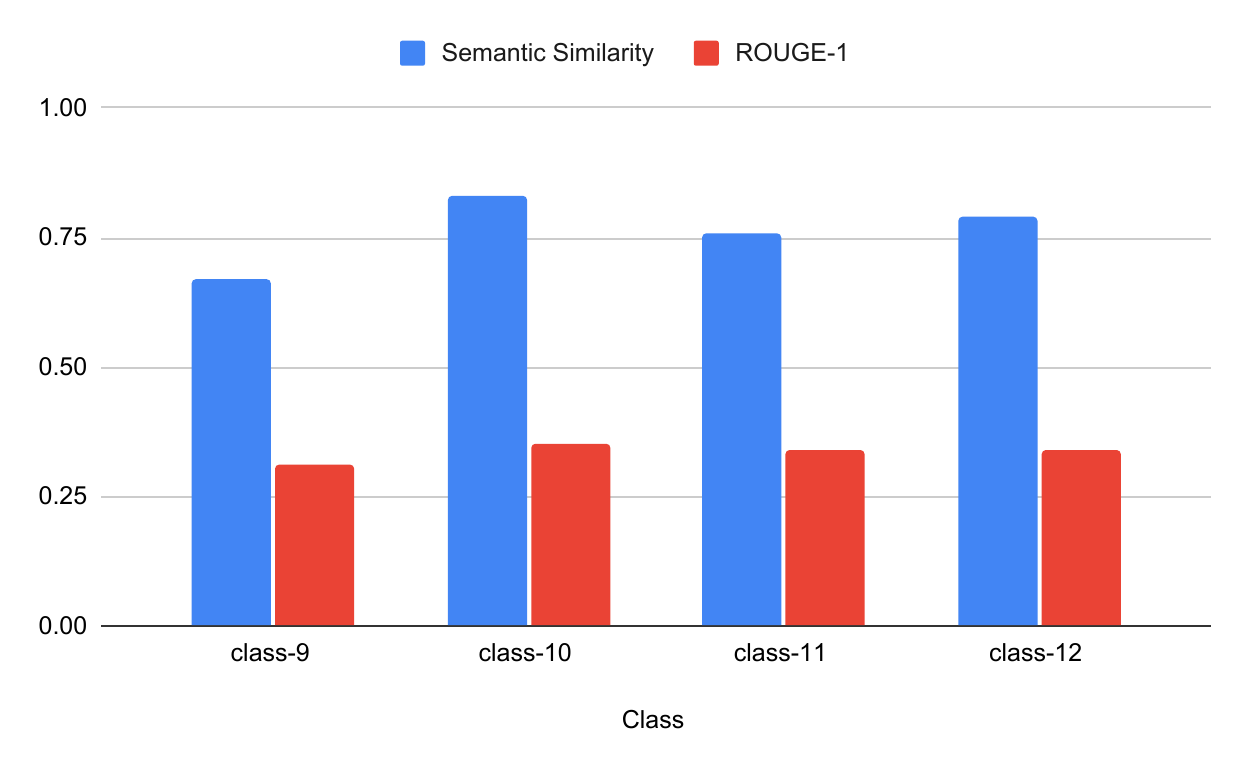}
        \label{fig:perf-class}
    }
    \hfill
    \subfigure[Ablation Study]{
        \includegraphics[width=0.45\textwidth]{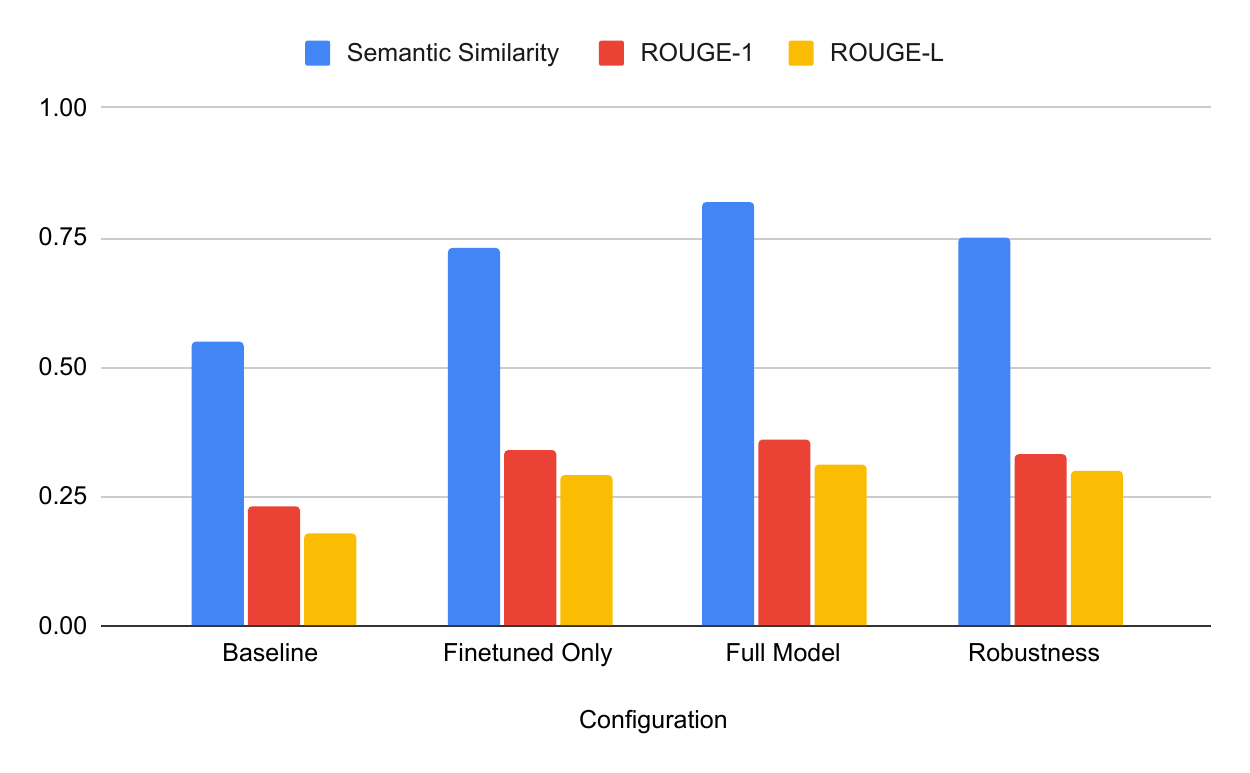}
        \label{fig:ablation}
    }

    \caption{Evaluation of GurukulAI's generated answers using ROUGE-L and Semantic Similarity metrics. Each subplot shows performance breakdown across (a) subject, (b) difficulty, (c) grade level, and (d) ablation study.}
    \label{fig:results}
\end{figure*}

We report two metrics:
\begin{itemize}
    \item \textbf{ROUGE-L:} Captures token-level overlap with ground-truth answers, useful for factual coverage~\cite{lin2004rouge}.
    \item \textbf{Semantic Similarity:} Computed as cosine similarity between Sentence-BERT embeddings of generated and reference answers, capturing contextual alignment.
\end{itemize}

\paragraph{Performance by Subject.}
Figure~\ref{fig:perf-subject} shows that GurukulAI performs best on Social Science and Science questions, achieving high scores in both ROUGE-L and semantic similarity. Performance on English and Hindi is slightly lower, likely due to the greater variability in acceptable answer phrasing for language subjects.

\paragraph{Performance by Class.}
As shown in Figure~\ref{fig:perf-class}, performance remains relatively stable across Classes 9–12, with a minor dip in Class 11. This could be attributed to the increase in question abstraction and complexity in senior grades.

\paragraph{Performance by Difficulty.}
Figure~\ref{fig:perf-difficulty} highlights a predictable trend: performance drops on “hard” questions, which require deeper reasoning and multi-sentence answers. While ROUGE-L is impacted more sharply, semantic similarity remains relatively robust, suggesting that even when answers are shorter or less token-aligned, they are semantically relevant.

Overall, these results suggest that GurukulAI can generate factually accurate and contextually relevant answers across subjects and grade levels, while showing room for improvement in analytical and abstract question handling.

\subsection{Qualitative User Study}

To evaluate the pedagogical usefulness and practical impact of \textsc{GurukulAI}, we conducted a structured human study with 43 student participants spanning Classes 9 to 12, as well as undergraduate levels with past CBSE experience. Each participant interacted with the platform and completed a feedback survey using a 5-point Likert scale across multiple aspects.

Table~\ref{tab:human-eval} summarizes the key findings. A majority of users found the system factually reliable (86\%), helpful (84\%), and easy to understand (81\%). High satisfaction was also reported in terms of response time (88\%) and improved subject understanding (89\%). Notably, 77\% of users found GurukulAI less distracting than conventional online search, and 72\% preferred it over existing methods.

\begin{table}[h]
\centering
\begin{tabular}{p{0.55\linewidth} c}
\toprule
\textbf{Evaluation Aspect} & \textbf{Agreement (\%)} \\
\midrule
Accurate and helpful responses & 84\% \\
Factually aligned with NCERT & 86\% \\
Explanations easy to understand & 81\% \\
Response time satisfaction & 88\% \\
Less distracting than online search & 77\% \\
Preferred over existing methods & 72\% \\
Improved subject understanding & 89\% \\
\bottomrule
\end{tabular}
\caption{Summary of user feedback on GurukulAI (N = 43).}
\label{tab:human-eval}
\end{table}

Open-ended responses praised the platform's user interface, speed, and textbook alignment. Participants described GurukulAI as “a personal tutor aligned with our books.” Suggestions for improvement included simplifying certain explanations, enhancing UI/UX consistency, and adding interactive follow-up features. While some preferred broader LLMs like ChatGPT or Gemini for general queries, most participants favored GurukulAI for its syllabus coverage and NCERT grounding.

These results suggest that \textsc{GurukulAI} provides a promising alternative to general-purpose LLMs for structured, curriculum-aligned educational support in the Indian context.

\subsection{Ablation Study}

We compare four model configurations on our test set to measure the effect of fine-tuning and retrieval:

\begin{enumerate}
    \item \textbf{Baseline:} Pretrained LLaMA 3.1 8B with no fine-tuning.
    \item \textbf{Fine-tuned Only:} Model trained on our QA dataset, without RAG.
    \item \textbf{Full Model (RAG):} Fine-tuned model with FAISS-based semantic retrieval.
    \item \textbf{Robustness:} Full model tested on noisy/incomplete queries.
\end{enumerate}

As shown in Figure~\ref{fig:ablation}, the full RAG model achieves the highest scores. Removing retrieval reduces Semantic similarity by 13\% and ROUGE-L by 6--8\%, confirming the benefit of chapter-level grounding.

% \begin{figure}[h]
%     \centering
%     \includegraphics[width=0.95\linewidth, trim={0 0 0 0}, clip]{latex/fig/result4.pdf}
%     \caption{Performance comparison across model variants. RAG improves semantic alignment and factual coverage.}
%     \label{fig:ablation}
% \end{figure}

\section{Conclusion and Future Work}

\textit{GurukulAI} is a curriculum-aligned, AI system designed for the Indian school education context. It integrates a fine-tuned generative model, a retrieval-augmented inference pipeline, and a syllabus-grounded QA dataset to deliver accurate, chapter-specific answers and adaptive learning support. Unlike generic tools, it enables MCQ and theory evaluation through a public web interface grounded in NCERT content.

In future work, we plan to expand subject and class coverage beyond Grades 9–12, support regional board syllabi, and enhance multimodal capabilities (e.g., diagrams and equations). Currently, analytical subjects such as Mathematics and Physics are excluded from the portal, and additional instruction tuning on domain-specific datasets is ongoing. We also aim to implement adaptive learning paths and lightweight user modeling to personalize learning. Our evaluations demonstrate the feasibility of scalable, culturally grounded generative AI for inclusive education in India.

\section{Limitations}

Despite encouraging results, several limitations remain. Currently, \textsc{GurukulAI} supports only Classes 9–12 (science stream) and is aligned solely with NCERT textbooks, limiting applicability to other boards such as ICSE or state curricula.

The model struggles with high-cognitive-load questions requiring multi-step reasoning, especially in subjects like Mathematics and Physics. It also lacks multimodal capabilities, and cannot process visual inputs such as diagrams or charts.

As a black-box generative model, it may occasionally produce factually incorrect or misleading answers ("hallucinations"), despite contextual retrieval. Users are advised to cross-verify responses before relying on them for academic use.

These limitations indicate future work directions: broader curriculum support, improved reasoning, multimodal input handling, and more interpretable model behavior.

\section{Ethics}

We followed ethical practices during the development of \textsc{GurukulAI}. The QA dataset was curated from publicly available NCERT textbooks and educational websites under fair-use for non-commercial academic purposes. No personal or user data was collected during training or evaluation.

To ensure quality, dataset entries were filtered using the Claude Sonnet API and manually reviewed. Some biases may persist due to automated processing. The system currently supports only English and Hindi, which may limit accessibility for other language users.

\textsc{GurukulAI} is intended as a supplementary learning tool, not a substitute for formal education. Users are advised to verify critical responses.

\bibliography{custom}

@article{wang2023survey,
  title={A Survey on Large Language Models for Education: Applications, Challenges, and Opportunities},
  author={Wang, Xiaohan and Chen, Yi and others},
  journal={arXiv preprint arXiv:2306.11795},
  year={2023}
}

@article{khan2023survey,
  title={A Systematic Review of ChatGPT in Education: Opportunities, Challenges, and Ethical Concerns},
  author={Khan, Rubina and Alotaibi, Nada and others},
  journal={Education and Information Technologies},
  year={2023}
}

@article{gilardi2023chatgpt,
  title={ChatGPT Out of the Box: How Do LLMs Generalize to Non-Western Contexts?},
  author={Gilardi, Fabrizio and Gessler, Timo and Kubli, Mirko},
  journal={arXiv preprint arXiv:2302.06598},
  year={2023}
}

@inproceedings{kakwani2020indicnlp,
  title={IndicNLPSuite: Monolingual Corpora, Evaluation Benchmarks and Pre-trained Multilingual Language Models for Indian Languages},
  author={Kakwani, Divyanshu and Khanuja, Simran and others},
  booktitle={Findings of EMNLP},
  year={2020}
}

@misc{udise2023,
  title = {UDISE+ 2023-24 Report Summary},
  author = {{Ministry of Education, Government of India}},
  year = {2024},
  url = {https://www.drishtiias.com/daily-updates/daily-news-analysis/udise-report-2023-24},
  note = {Accessed: 2025-06-30}
}

@inproceedings{rajpurkar2016squad,
  title={SQuAD: 100,000+ Questions for Machine Comprehension of Text},
  author={Rajpurkar, Pranav and Zhang, Jian and Lopyrev, Konstantin and Liang, Percy},
  booktitle={EMNLP},
  year={2016}
}

@inproceedings{yang2018hotpotqa,
  title={HotpotQA: A Dataset for Diverse, Explainable Multi-hop Question Answering},
  author={Yang, Zhilin and Qi, Peng and Zhang, Saizheng and Bengio, Yoshua and Cohen, William W. and Salakhutdinov, Ruslan and Manning, Christopher D.},
  booktitle={EMNLP},
  year={2018}
}

@inproceedings{choi2020ednet,
  title={EdNet: A Large-Scale Hierarchical Dataset in Education},
  author={Choi, Yoonho and Lee, Seewoo and Shin, Dongmin and Heo, Beomsu and Jang, Yanghoon and Oh, Alice and Shin, Jinwoo},
  booktitle={AIED},
  year={2020}
}

@inproceedings{garg2023iexamqa,
  title={iExamQA: A Dataset for Multilingual and Multimodal Question Answering in Indian Exams},
  author={Garg, Shikhar and Bhardwaj, Kritika and others},
  booktitle={Findings of ACL},
  year={2023}
}

@article{wang2023llms,
  title={A survey on large language models for education: Applications, challenges, and opportunities},
  author={Wang, Xiaohan and Chen, Yi and others},
  journal={arXiv preprint arXiv:2306.11795},
  year={2023}
}

@article{khan2023chatgpt,
  title={A systematic review of ChatGPT in education: Opportunities, challenges, and ethical concerns},
  author={Khan, Rubina and Alotaibi, Nada and others},
  journal={Education and Information Technologies},
  year={2023}
}

@inproceedings{holstein2018classroom,
  title={Classroom orchestration: Challenges and opportunities for learning at scale},
  author={Holstein, Kenneth and McLaren, Bruce M and Aleven, Vincent},
  booktitle={Proceedings of the Fifth Annual ACM Conference on Learning at Scale},
  year={2018}
}

@article{gilardi2023nonwestern,
  title={ChatGPT out of the box: How do LLMs generalize to non-western contexts?},
  author={Gilardi, Fabrizio and Gessler, Timo and Kubli, Mirko},
  journal={arXiv preprint arXiv:2302.06598},
  year={2023}
}

@inproceedings{kakwani2020indic,
  title={IndicNLPSuite: Monolingual corpora, evaluation benchmarks and pre-trained multilingual language models for Indian languages},
  author={Kakwani, Divyanshu and Khanuja, Simran and others},
  booktitle={Findings of EMNLP},
  year={2020}
}

@inproceedings{dua2019drop,
  title={DROP: A reading comprehension benchmark requiring discrete reasoning over paragraphs},
  author={Dua, Dheeru and Wang, Yizhong and others},
  booktitle={Proceedings of NAACL},
  year={2019}
}

@inproceedings{lai2017race,
  title={RACE: Large-scale reading comprehension dataset from examinations},
  author={Lai, Guokun and Xie, Qizhe and others},
  booktitle={Proceedings of EMNLP},
  year={2017}
}

@misc{bge2023,
  title={BGE M3 Embeddings},
  author={BAAI Team},
  howpublished={\url{https://huggingface.co/BAAI/bge-m3}},
  year={2023}
}

@misc{faiss2017,
  title={FAISS: A library for efficient similarity search},
  author={Johnson, Jeff and Douze, Matthijs and Jégou, Hervé},
  howpublished={\url{https://github.com/facebookresearch/faiss}},
  year={2017}
}

@inproceedings{lin2004rouge,
  title={ROUGE: A package for automatic evaluation of summaries},
  author={Lin, Chin-Yew},
  booktitle={Text summarization branches out},
  pages={74--81},
  year={2004}
}

@inproceedings{singh2025indicqa,
  title={INDIC QA Benchmark: A Multilingual Benchmark to Evaluate Question Answering Capability of LLMs for Indic Languages},
  author={Singh, Saurabh and others},
  booktitle={Findings of NAACL},
  year={2025}
}

@misc{anthropic2023claude,
  author       = {Anthropic},
  title        = {Claude Language Model},
  howpublished = {\url{https://www.anthropic.com/index/claude}},
  year         = {2023},
  note         = {Accessed June 2025}
}

@misc{pymupdf2022,
  author       = {Rauber, Jorj and contributors},
  title        = {PyMuPDF: Python bindings for MuPDF (fitz)},
  year         = {2022},
  howpublished = {\url{https://github.com/pymupdf/PyMuPDF}},
  note         = {Version 1.22, Accessed June 2025}
}

@misc{selenium,
  title={SeleniumHQ Browser Automation},
  author={{Selenium}},
  year={2023},
  note={\url{https://www.selenium.dev/}}
}
\newpage
\appendix
\newpage

\end{document}